\documentclass{article}

\usepackage[final,nonatbib]{nips_2018}

\usepackage[utf8]{inputenc} 
\usepackage[T1]{fontenc}    
\usepackage{hyperref}       
\usepackage{url}            
\usepackage{booktabs}       
\usepackage{amsfonts}       
\usepackage{microtype}      
\usepackage{subfigure}
\usepackage{epsfig}
\usepackage{graphicx}
\usepackage{amsmath}
\usepackage{amssymb}
\usepackage{float}

\title{FermiNets: Learning generative machines to generate efficient neural networks via generative synthesis}

\author{
  Alexander Wong$^{1,2}$, Mohammad Javad Shafiee$^{1,2}$, Brendan Chwyl$^2$, and Francis Li$^2$\\
  $^{1}$Waterloo Artificial Intelligence Institute, University of Waterloo, Waterloo, ON, Canada\\
  $^{2}$DarwinAI Corp., Waterloo, ON, Canada\\
  \texttt{\{a28wong,mjshafiee\}@uwaterloo.ca, \{francis,brendan\}@darwinai.ca}
}

\begin{document}

\maketitle

\begin{abstract}
The tremendous potential exhibited by deep learning is often offset by architectural and computational complexity, making widespread deployment a challenge for edge scenarios such as mobile and other consumer devices.  To tackle this challenge, we explore the following idea: \textbf{Can we learn generative machines to automatically generate deep neural networks with efficient network architectures?}  In this study, we introduce the idea of \textbf{generative synthesis}, which is premised on the intricate interplay between a generator-inquisitor pair that work in tandem to garner insights and learn to generate highly efficient deep neural networks that best satisfies operational requirements.  What is most interesting is that, once a generator has been learned through generative synthesis, it can be used to generate not just one but a large variety of different, unique highly efficient deep neural networks that satisfy operational requirements.  Experimental results for image classification, semantic segmentation, and object detection tasks illustrate the efficacy of generative synthesis in producing generators that automatically generate highly efficient deep neural networks (which we nickname \textbf{FermiNets}) with higher model efficiency and lower computational costs (reaching >10$\times$ more efficient and fewer multiply-accumulate operations than several tested state-of-the-art networks), as well as higher energy efficiency (reaching >4$\times$ improvements in image inferences per joule consumed on a Nvidia Tegra X2 mobile processor).  As such, generative synthesis can be a powerful, generalized approach for accelerating and improving the building of deep neural networks for on-device edge scenarios.
\end{abstract}

\section{Introduction}
\vspace{-0.15in}
Deep learning~\cite{lecun2015deep} has garnered tremendous attention, driven by recent breakthroughs in a wide range of applications such as image categorization~\cite{krizhevsky2012imagenet} and speech recognition~\cite{DeepSpeech}.  Despite these successes, the increasing complexities of deep neural networks limit their widespread adoption in edge scenarios such as mobile and other consumer devices where computational, memory, bandwidth, and energy resources are scarce.  There has in recent years been a noticeable increase in the exploration of strategies to improve the efficiency of deep neural networks, particularly on edge and mobile devices. One common approach explored is precision reduction~\cite{Jacob,Meng2017,Courbariaux2015}, in which the data representation of a network is reduced from typical 32-bit floating point precision to fixed-point or integer precision~\cite{Jacob}, 2-bit precision~\cite{Meng2017}, or even 1-bit precision~\cite{Courbariaux2015}.   Another strategy explored is model compression~\cite{han2015deep,distillation,projectionnet}, which involves traditional data compression methods such as weight thresholding, hashing, and Huffman coding, as well as teacher-student strategies involving a larger teacher network training a smaller student network.  Finally, another technique that has been investigated targets the fundamental design of deep neural networks and involves leveraging architectural design principles to achieve more efficient deep neural network macroarchitectures~\cite{MobileNet,SqueezeNet,Squishednet,ShuffleNet}.

In this study, we explore a very different idea: \textbf{Can we learn generative machines to automatically generate deep neural networks with efficient network architectures?}  We introduce the idea of \textbf{generative synthesis}, which is premised on the intricate interplay between a generator-inquisitor pair that work in tandem to garner insights and learn to generate highly efficient deep neural networks that best satisfies operational requirements, such as those needed for on-device edge scenarios.
\vspace{-0.1in}
\section{Generative Synthesis}
\vspace{-0.15in}
For the sake of brevity, a brief mathematical description of the methodology behind generative synthesis is provided as follows. Let a deep neural network be defined as a graph $N=\left\{V,E\right\}$, comprising of a set $V$ of vertices $v \in V$ and a set $E$ of edges $e \in E$ that form the network.  Now, let a generator $\mathcal{G}(s;\theta_\mathcal{G})$ be defined as a function parameterized by $\theta_\mathcal{G}$ that, given a seed $s  \in S$, generates a deep neural network $N_s=\left\{V_s,E_s\right\}$ (i.e., $N_s = \mathcal{G}(s)$), where $S$ is the set of possible seeds.  Given operational requirements (characterized by indicator function $1_r(\cdot)$), the goal of generative synthesis is to learn an expression of $\mathcal{G}$ that can generate deep neural networks $\left\{N_s|s \in S\right\}$ that maximize a universal performance function $\mathcal{U}$ (e.g.,~\cite{NetScore}) while satisfying requirements defined by $1_r(\cdot)$.  One can thus formulate the problem of learning $\mathcal{G}$ as the following constrained optimization problem,
\vspace{-0.05in}
\begin{equation}
\mathcal{G}  = \max_{\mathcal{G}}~\mathcal{U}(\mathcal{G}(s))~~\textrm{subject~to}~~1_r(\mathcal{G}(s))=1,~~\forall s \in S.
\label{optimization}
\end{equation}
\noindent Solving the problem posed in Eq.~\ref{optimization} is highly non-trivial given the enormity of the feasible region, and thus an efficient approach for solving this optimization problem is highly desired.

In {\it generative synthesis}, we find an approximate solution to the problem posed in Eq.~\ref{optimization} by leveraging the interplay between a generator-inquisitor pair $\left\{\mathcal{G},\mathcal{I}\right\}$, with $\mathcal{G}$ denoting a generator and $\mathcal{I}$ denoting an inquisitor that work in tandem to obtain improved insights about deep neural networks as well as learn to generate highly efficient networks in a cyclical manner.  More specifically, let $\mathcal{I}(\mathcal{G};\theta_\mathcal{I})$ be defined as a function parameterized by $\theta_\mathcal{I}$ that, given a generator $\mathcal{G}$, produces a set of parameter changes $\Delta\theta_\mathcal{G}$ (i.e., $\Delta\theta_\mathcal{G} = \mathcal{I}(\mathcal{G})$).
At initialization, both $\theta_{\mathcal{G}}$ and $\theta_{\mathcal{I}}$ are initialized based on prototype $\varphi$, $\mathcal{U}$, and $1_r(\cdot)$, resulting in $\mathcal{G}_0$ and $\mathcal{I}_0$.  After initialization, at each cycle $k$, the generator at cycle $k$ (i.e., $\mathcal{G}_k$) generates a new deep neural network ${N}_{s^k}$ based on a generated seed $s_k$ (i.e., ${N}_{s^k} = \mathcal{G}_k\left(s_k\right)$).

After the generation of ${N}_{s^k}$ using $\mathcal{G}_k$, $\left\{\mathcal{V}_{s^k},\mathcal{E}_{s^k}\right\}$ in ${N}_{s^k}$ are probed with a set $X$ of targeted stimulus signals $x \in X$ and the corresponding set $Y_{\mathcal{G}_k\left(s_k\right)}$ of reactionary response signals $y \in Y_{\mathcal{G}_k\left(s_k\right)}$ are observed, where $\mathcal{V}_{s^k} \subseteq V_{s^k}$ and $\mathcal{E}_{s^k} \subseteq E_{s^k}$.  After the observation process, $\theta_{\mathcal{I}}$ is updated based on $Y_{\mathcal{G}_k\left(s_k\right)}$, $\mathcal{U}(\mathcal{G}_k\left(s_k\right))$, and $1_r(\mathcal{G}_k\left(s_k\right))$ to obtain $\mathcal{I}_{k+1}$.  In particular, by probing $\left\{\mathcal{V}_{s^k},\mathcal{E}_{s^k}\right\}$ in ${N}_{s^k}$ with $X$ and observing $Y_{\mathcal{G}_k\left(s_k\right)}$, the inquisitor $\mathcal{I}$ is able to learn at a foundational level about the architectural efficiencies of ${N}_{s^k}$ via information-theoretic insights that are derived from $Y_{\mathcal{G}_k\left(s_k\right)}$.

After the inquisitor update process, $\theta_{\mathcal{G}}$ is updated according to $\Delta\theta_{\mathcal{G}_{k}\left(s_k\right)}$ produced by $\mathcal{I}_{k+1}$ to obtain $\mathcal{G}_{k+1}$.  The aforementioned process of generating, probing, observation, and updating is repeated over cycles, resulting in a sequence of improving approximate solutions of $\mathcal{G}$ to the problem in Eq.~\ref{optimization}.  What is most interesting about generative synthesis is that, once a generator $\mathcal{G}$ has been learned, it can be used to generate not just one but a large variety of different, unique highly efficient deep neural networks ${N}$, using different seeds $s \in S$, that satisfy operational requirements defined by $1_r(\cdot)$.
\vspace{-0.1in}
\section{Experimental Results and Discussion}
\vspace{-0.15in}
To evaluate the efficacy of generative synthesis (which we will refer to as \textbf{GenSynth} for short from here on) in producing generators that automatically generate highly efficient deep neural networks (which we nickname \textbf{FermiNets} because of their very small sizes), three experiments were performed:
\vspace{-0.25in}
\begin{itemize}
\item  \textbf{Image classification}. $\varphi$: ResNet~\cite{ResNet}, $1_r(\cdot)$: accuracy on CIFAR-10 $\geq$ 89\%.
\item  \textbf{Semantic segmentation}. $\varphi$: RefineNet~\cite{RefineNet}, $1_r(\cdot)$: accuracy on CamVid~\cite{CamVid} $\geq$ 90\%.
\item \textbf{Object detection}. $\varphi$: DetectNet~\cite{DetectNet}, $1_r(\cdot)$: mAP on Parse27K~\cite{Parse27K} $\geq$ 61\%.
\end{itemize}
\vspace{-0.1in}
The performance of the generated FermiNets was evaluated using these metrics: i) \textbf{information density~\cite{Canziani}} as a metric for assessing model efficiency, ii) \textbf{multiply-accumulate (MAC) operations} as a metric for computational cost, and iii) \textbf{NetScore~\cite{NetScore}} as a metric for assessing overall network performance (balance between accuracy, architectural complexity, and computational complexity).

\begin{figure*}
	\centering
\vspace{-0.1in}
	\includegraphics[width = 0.45\linewidth]{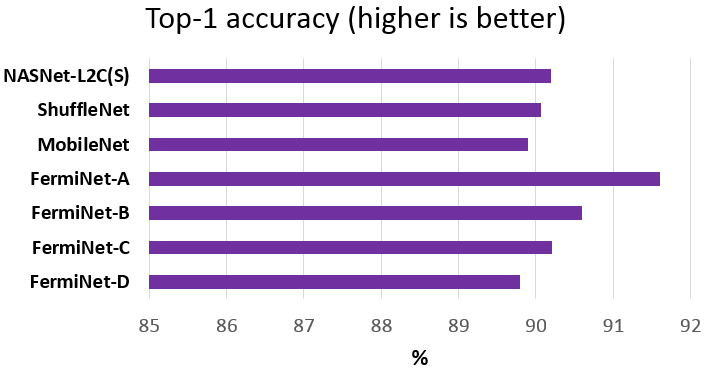}~\includegraphics[width = 0.45\linewidth]{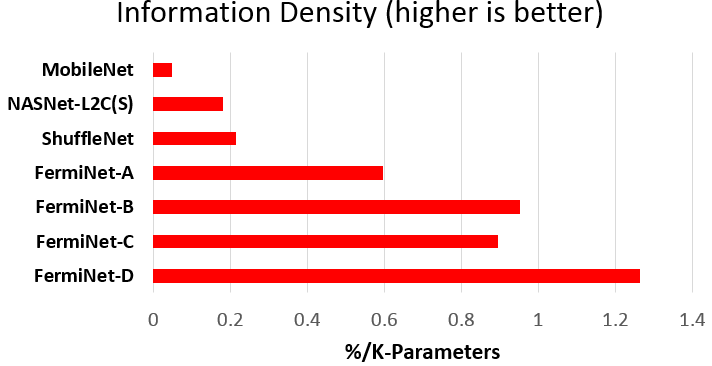}\\~\\
\vspace{-0.15in}
\includegraphics[width = 0.45\linewidth]{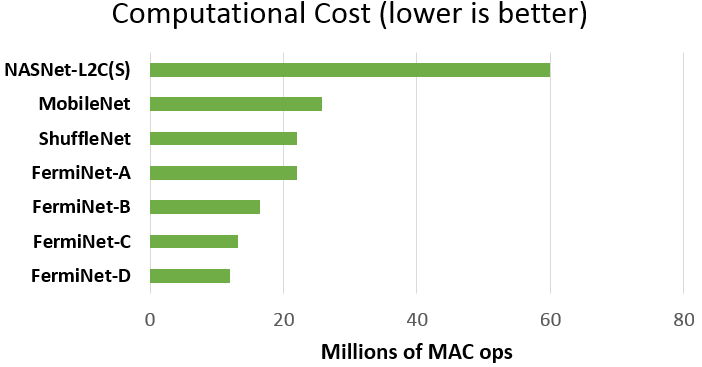}~\includegraphics[width = 0.45\linewidth]{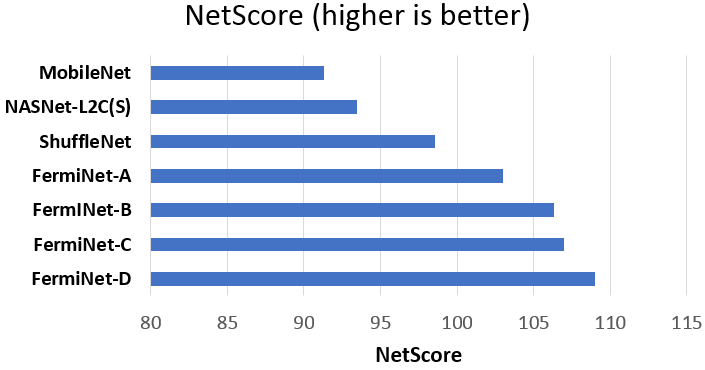}
\vspace{-0.14in}
	\caption{Image classification: (top left) Top-1 accuracy on CIFAR-10, (top right) Information density~\cite{Canziani}, (bottom left) MAC operations, and (bottom right) NetScore~\cite{NetScore}}
\vspace{-0.11in}
	\label{fig6}
\end{figure*}

\textbf{Image classification}: The top-1 test accuracy of the generated FermiNets along with MobileNet~\cite{MobileNet}, ShuffleNet~\cite{ShuffleNet}, and NASNet-L2C(S)~\cite{L2C} is shown in Fig.~\ref{fig6} (top left), with FermiNet-A, FermiNet-B, and FermiNet-C providing the highest top-1 accuracies amongst the tested state-of-the-art efficient networks ($\sim$1.4\%, $\sim$0.4\%, and $\sim$0.01\% higher than NASNet-L2C(S), respectively).  As shown in Fig.~\ref{fig6} (top right), the information densities of the generated FermiNets had noticeably higher information densities than MobileNet, ShuffleNet, as well as NASNet-L2C(S)\footnote{Note that, since the number of parameters for NASNet-L2C(S) was not reported in~\cite{L2C}, it was approximated based on the reported model size to enable the computation of information density and NetScore.} which was produced via a combination of state-of-the-art model compression approaches~\cite{Jacob,distillation,projectionnet}.  In particular, FermiNet-A, FermiNet-B, FermiNet-C, and FermiNet-D have information densities that exceeds that of MobileNet by \textbf{>12.5$\times$}, \textbf{$\sim$20$\times$}, \textbf{$\sim$19$\times$}, and \textbf{$>$26.4$\times$}, respectively.  As shown in Fig.~\ref{fig6} (bottom left), the FermiNets require noticeably fewer number of MAC operations than the other tested networks, with that of FermiNet-A, FermiNet-B, FermiNet-C, and FermiNet-D less than that of NASNet-L2C(S) by \textbf{>2.7$\times$}, \textbf{>3.6$\times$}, \textbf{>4.5$\times$}, and \textbf{$\sim$5$\times$}, respectively.  To evaluate overall network performance, the NetScore of the tested networks is shown in Fig.~\ref{fig6} (bottom right), with the FermiNets having noticeably higher NetScores when compared to MobileNet, ShuffleNet, and NASNet-L2C(S).  In particular, the NetScore of FermiNet-A, FermiNet-B, FermiNet-C, and FermiNet-D exceed that of MobileNet by \textbf{>11.8} points, \textbf{>15} points, \textbf{>15.6}, and \textbf{>17.5} points, respectively.  These significant improvements in NetScore achieved by the generated FermiNets illustrate the power of GenSynth in striking a strong balance between accuracy, architectural complexity, and computational cost, which is important for on-device edge scenarios.

\begin{figure*}[t]
	\hspace{-0.2in}\includegraphics[width = 0.35\linewidth]{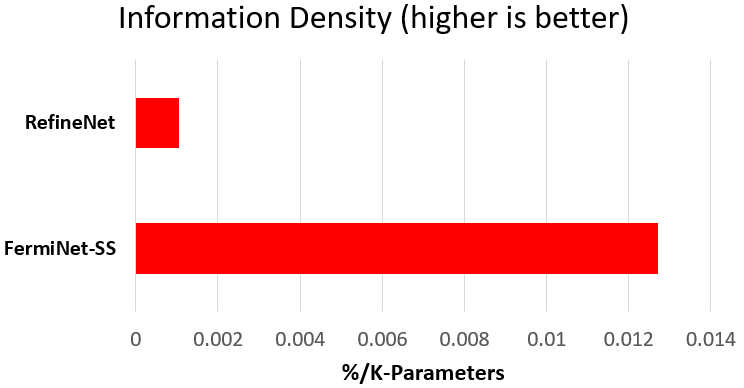}~\includegraphics[width = 0.35\linewidth]{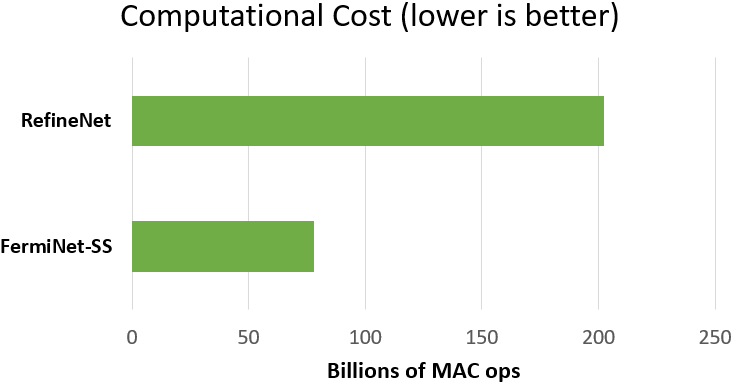}~
\includegraphics[width = 0.35\linewidth]{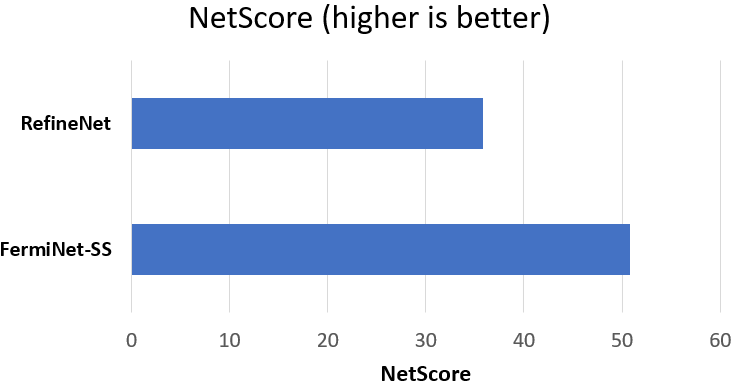}
\vspace{-0.2in}
	\caption{Semantic segmentation: (a) Information density, (b) MAC operations, and (c) NetScore.}
\vspace{-0.26in}
	\label{fig3}
\end{figure*}

\textbf{Semantic segmentation}: The pixel-wise accuracy of RefineNet~\cite{RefineNet} and the generated FermiNet-SS are 90.3\% and 90.4\%, respectively.  As shown in Fig.~\ref{fig3}(a), the information density of FermiNet-SS exceeds that of RefineNet by \textbf{$>$12$\times$}, thus illustrate its modelling efficiency.  As shown in Fig.~\ref{fig3}(b), the number of MAC operations used by FermiNet-SS is less than that of RefineNet by \textbf{$\sim$2.6$\times$}.  Finally, as shown in Fig.~\ref{fig3}(c), the NetScore of FermiNet-SS exceeds that of RefineNet by \textbf{$\sim$15} points.

\begin{figure*}[t]
	\hspace{-0.7in}\includegraphics[width = 0.32\linewidth]{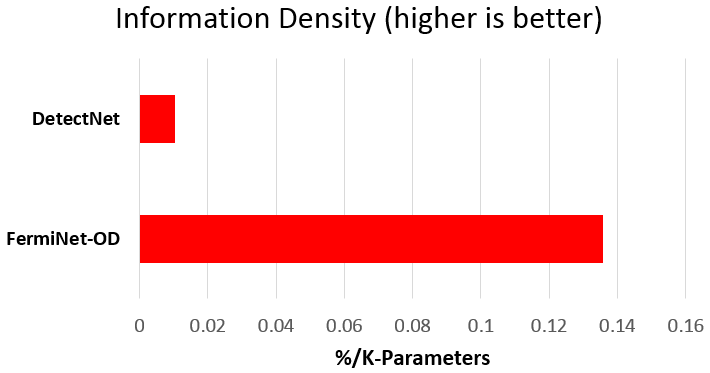}~\includegraphics[width = 0.32\linewidth]{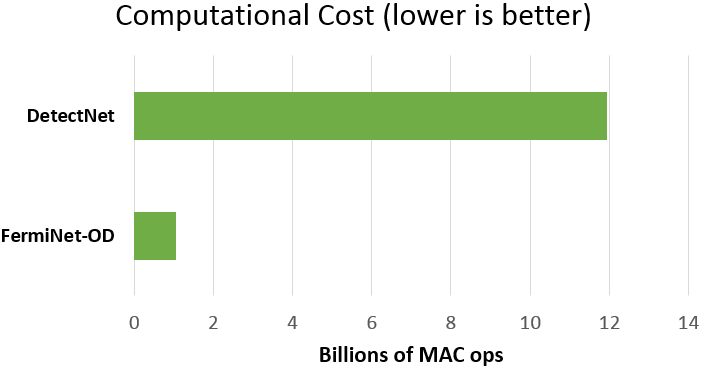}~
\includegraphics[width = 0.32\linewidth]{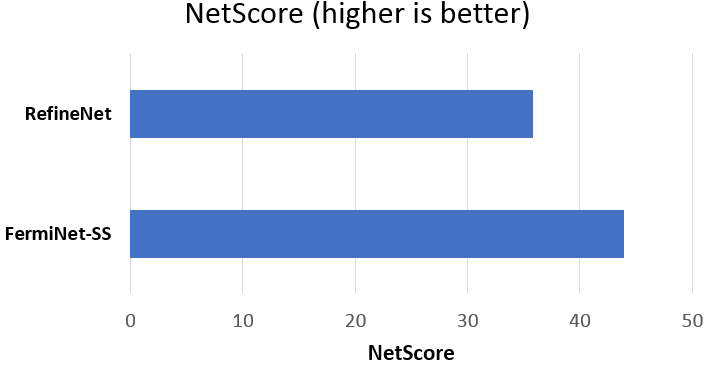}~\includegraphics[width = 0.32\linewidth]{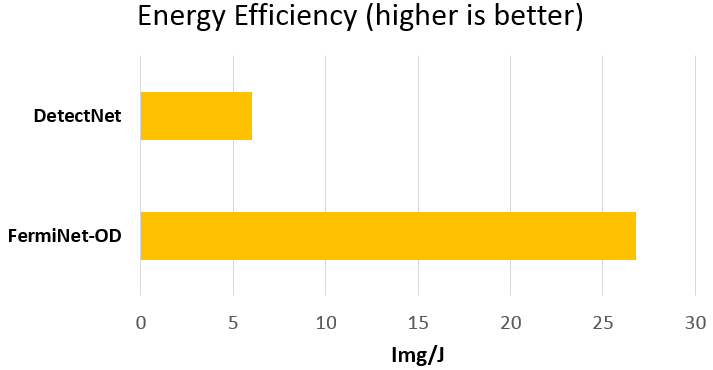}
\vspace{-0.15in}
	\caption{Object detection: (a) Information density, (b) MAC operations, (c) NetScore, and (d) energy efficiency on the Nvidia Tegra X2 mobile processor.}
\vspace{-0.23in}
	\label{fig2}
\end{figure*}

\textbf{Object detection}: The mean average precision (mAP) of DetectNet~\cite{DetectNet} and the generated FermiNet-OD are 61.8\% and 61.0\%, respectively.  As shown in Fig.~\ref{fig2}(a), the information density of FermiNet-OD exceeds that of DetectNet by \textbf{>10$\times$}.  The number of MAC operations used by FermiNet-OD along with that of DetectNet is shown in Fig.~\ref{fig2}(b).  It can be observed that the number of MAC operations used by FermiNet-OD less than that of DetectNet by \textbf{>11$\times$}.  As shown in Fig.~\ref{fig2}(c), the NetScore of FermiNet-OD exceeds that of DetectNet by \textbf{>21} points.  In an additional experiment, we study the energy efficiency of FermiNet-OD when operating in edge device scenarios.  To quantitatively assess the energy efficiency during inference, the metric we leverage in this study is the number of image inferences per joule consumed (img/J) by a Nvidia Tegra X2 mobile processor.  As shown in Fig.~\ref{fig2}(d), FermiNet-OD is significantly more energy efficient than DetectNet, enabling the Tegra X2 mobile processor to process \textbf{>4$\times$} more images per joule than can be achieved with DetectNet.

As shown by the empirical results in this study, it can be seen that GenSynth can be a powerful, generalized approach for building deep neural networks that satisfies operational requirements for on-device edge scenarios such as mobile and other consumer devices.
\vspace{-0.1in}
\section*{Acknowledgements}
\vspace{-0.15in}
The authors thank Akif Kamal, Michael St. Jules, Stanislav Bochkarev, David Dolson, Xiao Yu Wang, and Desmond Lin at DarwinAI Corp. for their support and assistance in experimental preparations.
\vspace{-0.1in}
{\footnotesize
\bibliographystyle{abbrv}
\vspace{-0.1in}
\bibliography{ccn_style}
}

\end{document}